\documentclass[letterpaper,10pt,conference]{ieeeconf}
\IEEEoverridecommandlockouts
\usepackage[utf8]{inputenc}
\usepackage[T1]{fontenc}
\usepackage{cite}
\usepackage{amsmath,amssymb,amsfonts}
\usepackage{graphicx}
\usepackage{textcomp}
\usepackage{url}
\usepackage{xcolor}
\usepackage{booktabs}
\usepackage{tabularx}
\usepackage{makecell}
\usepackage{multirow}
\usepackage{subcaption}
\usepackage{tikz}
\usetikzlibrary{arrows.meta, positioning, shapes.geometric}
\usepackage{listings}
\usepackage{caption}
\definecolor{codegray}{RGB}{245,245,245}
\definecolor{framegray}{RGB}{180,180,180}

\lstdefinestyle{systemprompt}{
    basicstyle=\ttfamily\small,
    backgroundcolor=\color{codegray},
    frame=single,
    rulecolor=\color{framegray},
    breaklines=true,
    breakatwhitespace=false,
    columns=fullflexible,
    keepspaces=true,
    showstringspaces=false,
    tabsize=2,
    captionpos=b
}

\begin{document}

\title{\LARGE \bf KINO: A Keyframe Interface for VLM Planning and \\
Whole-Body Control in Humanoid Loco-Manipulation}

\author{Sitong Chen, Fatemeh Zargarbashi, Jin Cheng, Tianxu An, Stelian Coros\\
{\small ETH Z\"urich, Switzerland}}

\maketitle
\thispagestyle{empty}
\pagestyle{empty}

\begin{abstract}

Humanoid loco-manipulation requires robots to interpret task instructions and scene semantics while executing coordinated whole-body motions.
We propose a hierarchical framework that uses motion keyframes as an intermediate representation between Vision-Language Model (VLM) planning and Reinforcement Learning (RL) control.
Each keyframe specifies a target whole-body robot pose and, when applicable, an object pose.
Given a language instruction, scene observations, and execution feedback, the VLM selects successive task-relevant keyframes from a predefined library.
The selected keyframes are retargeted to the current scene to account for object poses and dimensions.
A keyframe-conditioned whole-body policy then generates joint-level actions to reach these goals.
We introduce a saliency-based keyframe sampling strategy for low-level policy training that improves end-to-end task success rate from 44\% to 92\% when using sparse VLM keyframes.
We evaluate our framework on object pickup, transport, and placement tasks in simulation and on a Unitree G1 humanoid.
The system successfully performs both one- and two-handed manipulation and generalises to placement locations beyond the training reference data.

\end{abstract}

\section{Introduction}

Humanoid loco-manipulation in real-world environments remains challenging due to the diversity and complexity of tasks and objects.
Successful execution requires not only understanding task semantics and decomposing it into smaller subtasks, but also generating dynamically feasible whole-body motions.
Humans naturally combine high-level reasoning with fast low-level processes when solving complex tasks~\cite{kahneman2011thinking}.
Similarly, robotic loco-manipulation can be decomposed into two complementary levels: high-level task planning and low-level motion control.
Integrating these levels remains a central challenge for general-purpose humanoid loco-manipulation.

Recent developments in VLMs have demonstrated strong capabilities in language interpretation and visual understanding, as well as reasoning and task planning \cite{driess2023palm}.
In robotics, Vision-Language-Action (VLA) models can directly predict robot motor actions from images and instructions\cite{kim2024openvla, brohan2023rt}. However, learning such end-to-end mappings requires the collection of large and diverse robot-specific interaction data, which is expensive and time-consuming.
On the other hand, RL has emerged as a powerful tool for robust low-level control of complex legged systems. RL can benefit from motion data obtained from retargeting human demonstrations\cite{yang2025omniretarget} to learn various motor skills such as walking and dancing.
Such policies are effective at following motions and accomplishing single tasks\cite{chen2025gmt, weng2025hdmi}.
However, scaling end-to-end RL to long-horizon tasks that require planning and transitioning among multiple skills remains challenging.

\begin{figure}[t]
    \centering
    \includegraphics[width=\linewidth]{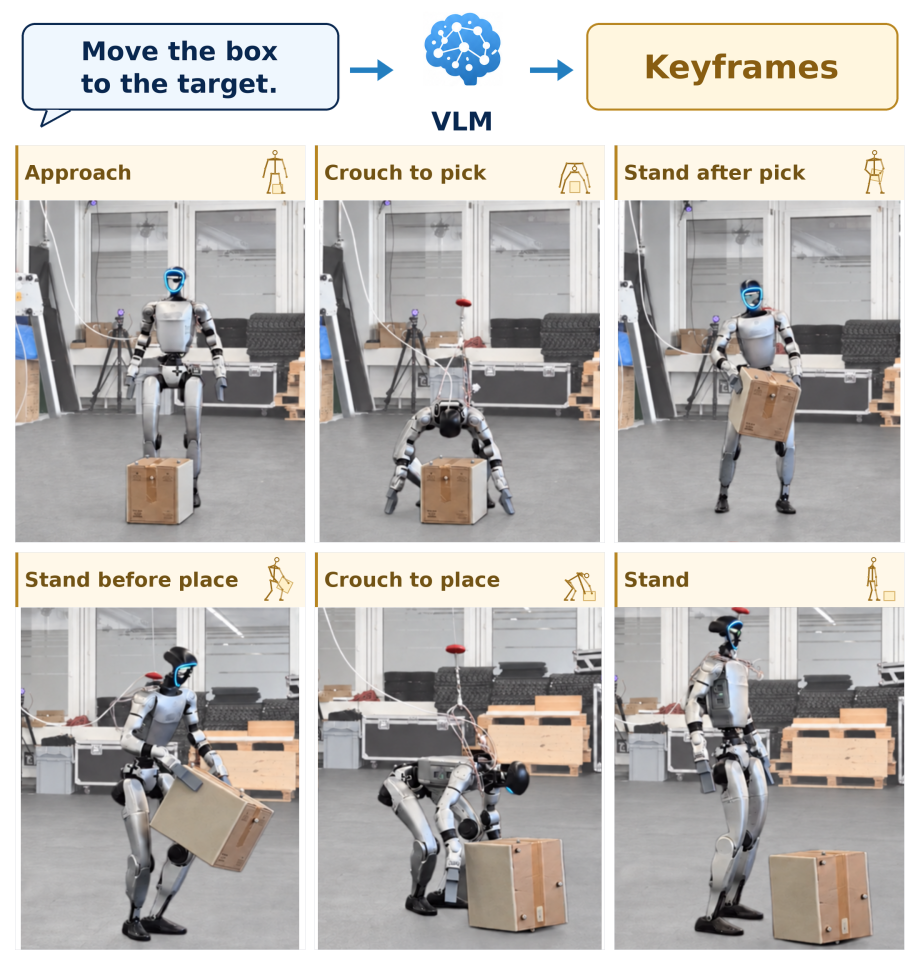}
    \caption{Our framework enables loco-manipulation tasks with humanoid robots in the real world.}
    \label{fig:teaser}
\end{figure}

Hierarchical approaches offer a promising direction to combine low-level RL policies for whole-body control and VLMs for high-level task reasoning.
Nevertheless, an important interface problem exists.
VLMs reason about semantic task objectives using vision and language, whereas low-level RL controllers typically require precise state observations and structured goal inputs~\cite{portela2025wholebody}, rather than open-ended language commands.

We aim to bridge this gap by using \textit{keyframes} as an intermediate representation between language-level planning and joint-level control.
To this end, we introduce KINO, a framework that integrates VLM-based reasoning and low-level whole-body control through keyframes (Fig. \ref{fig:teaser}).
Given a task command, scene observation and robot context, the VLM selects the next task-relevant keyframe from a pre-constructed keyframe library. The keyframe is then adjusted based on the current scene and provided as a goal to a keyframe-conditioned RL policy, which generates whole-body robot actions.

We evaluate our framework on a G1 humanoid robot both in simulation and the real world across one- and two-handed object pickup and transport tasks.
Our experiments show that KINO is able to successfully reason about subtasks required to accomplish the task, complete the task by following the keyframes, and generalise to placement regions beyond those represented in the training reference data.
We further show that saliency-based keyframe sampling improves end-to-end success from 44\% to 92\% under sparse keyframe goals.
These results demonstrate the potential of keyframes as an effective interface between semantic VLM planning and whole-body RL control.

Our main contributions are:
\begin{itemize}
    \item We introduce {KINO}, a hierarchical framework that bridges VLM-based semantic task planning and whole-body RL control using motion keyframes as a compact intermediate representation.
  \item We propose a saliency-based keyframe sampling for training low-level policies to reliably reach sparse keyframe goals provided by the VLM at inference time.
  \item We demonstrate that our framework has a high success rate in completing loco-manipulation tasks through simulation and hardware experiments.

\end{itemize}

\section{Related Work}

\subsection{Humanoid Loco-Manipulation}

Humanoid loco-manipulation requires coordinating locomotion, balance, whole-body control, and object interaction.
Recent reinforcement learning methods have enabled robust and diverse legged locomotion, from quadrupeds traversing difficult terrains \cite{lee2020learning, miki2022learning} to agile humanoid locomotion \cite{radosavovic2024real, cheng2024expressive,ji2024exbody2,he2025asap}.
For humanoids, their skeletal similarity to the human body provides an additional source of supervision: human motion can be used directly to specify desired behaviours, reducing the need for careful reward engineering.
This has motivated a line of work that leverages human motion through imitation. General motion-tracking policies can reproduce diverse behaviours by following kinematic full-body reference trajectories \cite{chen2025gmt,luo2026sonic,peng2025mimickit,zeng2025behavior}. Several methods use teleoperation systems to acquire robot skills for various tasks\cite{fu2024humanplus,he2024omnih2o,he2024learning}, while others generate dynamically feasible humanoid interaction data or learn physical interaction skills from human demonstrations through retargeting \cite{liu2025opt2skill,li2025learning}.
Recent work extends this paradigm to loco-manipulation.
HDMI \cite{weng2025hdmi} learns human-object interaction skills from video demonstrations, while OmniRetarget\cite{yang2025omniretarget} generates interaction-preserving kinematic references and then trains a tracker to imitate those motions.
ResMimic \cite{zhao2025resmimic} uses residual learning to adapt general motion-tracking policies to precise whole-body loco-manipulation.
SoftMimic \cite{margolis2025softmimic} modifies the reference motion based on a compliance factor to achieve safer interaction with the environment.
Despite their strong motion quality, these methods require a complete kinematic reference motion to specify the desired behaviour at inference time, making it difficult to be directly applicable in real-world scenarios.
In contrast, our approach requires only a sparse keyframe to be specified by the high-level planner without requiring to provide a complete reference trajectory.

\subsection{Foundation Models for Robot Planning and Control}

Recent VLMs have enabled natural
  language interpretation and reasoning about visual observations.
  VLA models extend these capabilities to control by mapping language and visual observations directly to robot actions \cite{brohan2023rt,kim2024openvla,black2024pi_0,intelligence2025pi}.
   Most existing VLA systems have been developed primarily for manipulation or navigation \cite{cheng2024navila}.
   Extending them to dynamic humanoid control is challenging because whole-body actions are high-dimensional and must be generated at high frequencies to maintain balance and manage contacts.
  Moreover, although VLA models benefit from large-scale vision-language pretraining,
  learning to output robot controls still requires large datasets of
  action-labelled robot trajectories. These data are expensive to collect and are often specific to a robot embodiment.

An alternative is to use foundation models as high-level reasoning modules rather than direct motor controllers.
In manipulation, prior work has used VLMs to modify structured programs \cite{liang2023code}, generate spatial representations of the environment \cite{huang2023voxposer}, or predict optimal actions through model predictive control \cite{zhao2024vlmpc}.
In legged robotics, FALCON \cite{he2025falcon} uses a VLM to coordinate locomotion and manipulation skills.
More closely related to our work, ReKep \cite{huang2024rekep} represents manipulation tasks through spatio-temporal relational keypoint constraints. However, it focuses on robot manipulation and does not address dynamic whole-body humanoid control.
Our work follows a hierarchical design in which the VLM selects the next keyframe based on the task and visual observation. This provides an explicit goal to be executed by a low-level policy. This design uses the reasoning capability of VLMs without requiring the extensive robot-specific data needed to learn end-to-end VLA models.

\subsection{Motion Keyframes}

A set of intermediate keyframes can represent diverse behaviours without prescribing complete trajectories, making them a general interface for motion control.
In animation, keyframes have been used as sparse constraints for controllable motion generation \cite{mo2023continuous,wang2026motionbricks}.
RobotKeyframing has used keyframes as high-level targets for robot locomotion policies, enabling robots to generate novel motions beyond those in the training data\cite{zargarbashi2024robotkeyframing}.
However, their method relies on manually specified keyframes at inference time.
Our approach provides a universal and compact interface to the low-level robot policy by using a foundation model to predict the next keyframe for a task.

\begin{figure*}[t]
    \centering
    \includegraphics[width=1\linewidth]{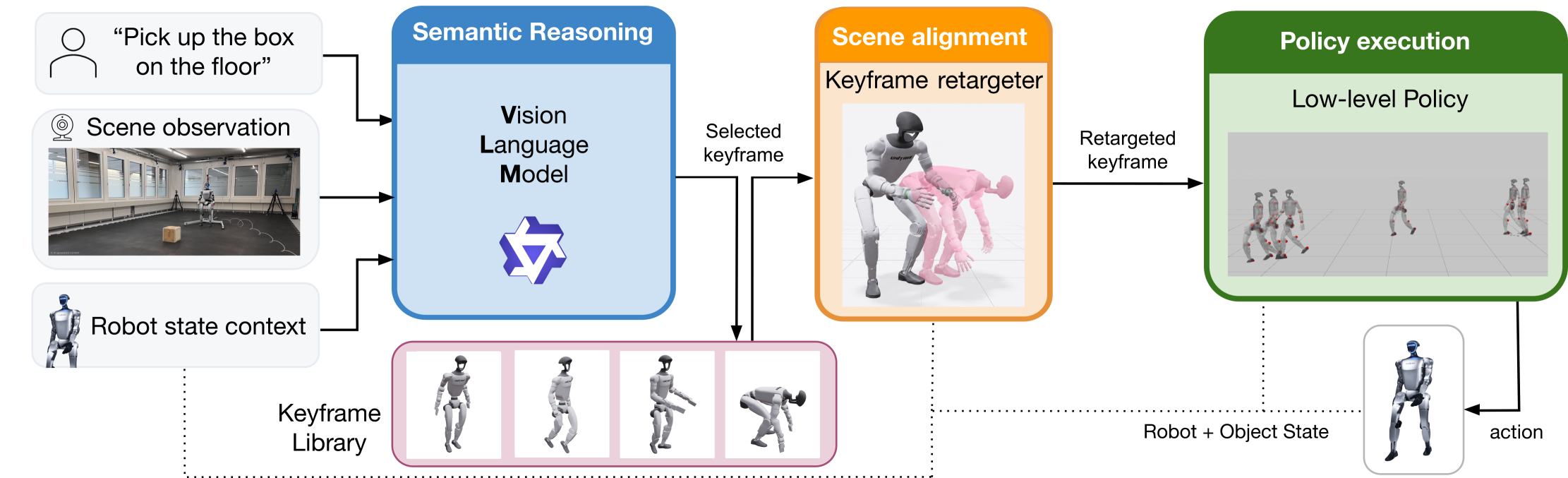}
    \caption{Overview of the pipeline. The VLM chooses a keyframe from a library based on the scene and robot context. The keyframe is then retargeted and passed to the low-level policy to complete the task.}
    \label{fig:system}
\end{figure*}

\section{Method}

In this work, we propose a hierarchical keyframe-based framework for humanoid loco-manipulation, as shown in Fig. \ref{fig:system}. The system decouples high-level semantic reasoning from low-level whole-body motion execution.
Given a text instruction, the current robot state context, and a visual observation of the scene, a VLM selects an appropriate keyframe from a pre-defined keyframe library.
Each keyframe represents a task-relevant intermediate pose, such as approaching, crouching, or picking an object.
This representation gives the system a compact and generalisable interface.
By restricting the VLM to keyframe selection rather than prediction of continuous joint commands, we eliminate the need for robot-specific data collection for high-level planning.
After a keyframe is selected, it is retargeted to the current scene by aligning it with the humanoid robot and the target object according to the task-specific spatial constraints. The retargeted keyframe is then passed to a low-level robot policy, which generates joint actions to move the robot towards the desired pose.

\subsection{Keyframe-Conditioned Policy}

We formulate low-level motion execution as a goal-conditioned RL problem. The policy receives a keyframe goal and produces whole-body actions to drive the humanoid towards this pose.
At each control step $t$, the policy receives proprioceptive robot state, object state, previous action, and the next keyframe. The observation vector can be written as:
\begin{equation}
    \boldsymbol{x}^\pi_t =
    \Big[
    \underbrace{\boldsymbol{v}^{r}_{t},\boldsymbol{\omega}^{r}_{t},\boldsymbol{g}^r_t,
    \boldsymbol{q}^r_t,\dot{\boldsymbol{q}}^r_t}_{\text{robot state}},
    \underbrace{\boldsymbol{p}^{o}_t,\boldsymbol{Q}^{o}_t}_{\text{object state}},
    \underbrace{\boldsymbol{a}_{t-1}}_{\text{prev. action}}, \underbrace{\mathcal{K}_t}_{\text{keyframe}}
    \Big],
\end{equation}
where $\boldsymbol{v}^{r}_{t}$ and $\boldsymbol{\omega}^{r}_{t}$ are the local linear and angular velocities of the robot base, $\boldsymbol{g}^{r}_t$ is the projected gravity vector, $\boldsymbol{q}^r_t$ and $\dot{\boldsymbol{q}}^r_t$ are the joint positions and velocities, respectively. The object pose is represented by its local position $\boldsymbol{p}^{o}_{t}$ and quaternion $\boldsymbol{Q}^{o}_{t}$.
$\mathcal{K}_t$ is the keyframe goal
represented as
\begin{equation}
    \mathcal{K}_t =
    \left(
    \boldsymbol{\hat{p}}^{r}_t,
    \boldsymbol{\hat{R}}^{r}_t,
    \boldsymbol{\hat{q}}^r_t,
    \boldsymbol{\hat{p}}^{o}_t,
    \boldsymbol{\hat{Q}}^{o}_t
    \right),
\end{equation}
where $\boldsymbol{\hat{p}}^{r}_t$ and $\boldsymbol{\hat{R}}^{r}_t$ denote the target root position and orientation in the base frame, $\boldsymbol{\hat{q}}_t$ denotes the target joint angles, and $\boldsymbol{\hat{p}}^{o}_t$ and $\boldsymbol{\hat{Q}}^{o}_t$ denote the target object position and orientation in the base frame, respectively.
If no object is intended in the keyframe, $\boldsymbol{\hat{p}}^{o}_t$ and $\boldsymbol{\hat{Q}}^{o}_t$ are set to zero.
The policy input uses a history of past 10 observations.

We use an asymmetric actor-critic where the critic receives additional privileged observations including reference root, joint, body, and object states, as well as goal-state information and goal-relative errors:
\begin{equation}
    \boldsymbol{x}^{V}_t =
    \Big[
    \boldsymbol{x}^{\pi}_t,
    \boldsymbol{y}^{\mathrm{ref}}_t,
    \Delta \mathcal{K}_t,
    \tau_t
    \Big],
\end{equation}
where $\boldsymbol{y}^{\mathrm{ref}}_t$ denotes the reference motion state, $\Delta \mathcal{K}_t$ the error between the current state and the goal, and $\tau_t$ the remaining time to the goal. The critic also uses a temporal history of 3 observations.
This privileged information facilitates value estimation during
  training, while the deployed policy relies only on observations readily available in the real world.

The policy is trained with PPO \cite{schulman2017proximal} and a reference motion database
including locomotion, box manipulation, and bucket manipulation
The rewards are divided into three groups: reference tracking $r^{\text{track}}_t$, keyframe goal reaching $r^{\text{goal}}_t$ and regularisation $r^{\text{reg}}_t$.
The tracking terms follow a DeepMimic formulation\cite{peng2018deepmimic} and encourage the robot to match the reference root pose and velocity, joint configuration, body poses, and object pose if available.
The keyframe reward encourages the robot to reach the selected keyframe and is only active at the target time, similar to \cite{zargarbashi2024robotkeyframing}.
Regularisation terms penalise torques, action rates, joint accelerations, and foot slipping.
To mitigate negative interference between different reward groups, we use a multi-critic formulation \cite{zargarbashi2024robotkeyframing}, where different reward groups are assigned separate value functions ($V^{\mathrm{track}}, V^{\mathrm{goal}}, V^{\mathrm{reg}}$).

The policy outputs an action vector, $\boldsymbol{a}_t \in \mathbb{R}^{29}$, corresponding to target joint positions of the Unitree G1 humanoid which are tracked by a PD controller.
To bridge the sim2real gap, we employ domain randomisation during training. This includes randomising friction, object and robot mass, centre-of-mass offsets, actuator gains, joint parameters, actuator delay, observation noise, and external pushes.

\subsection{Robot Keyframes}

\subsubsection{Keyframe Sampling for Policy Training}
\label{sec: keyframe selection}
We take representative frames in the reference motion as keyframe goals during training.
Since keyframes provide the interface between the high-level planner and the control policy, the distribution of training keyframes should resemble the keyframes predicted by a VLM at inference time.
For locomotion, randomly sampling future frames as keyframes can provide diverse and valid locomotion goals.
In contrast,
loco-manipulation clips such as pick-and-place involve distinct stages, such as approaching, lifting, transporting, and placing an object.
Uniform sampling may underrepresent these events relative to the many intermediate frames of approaching or walking, and therefore may fail to expose the policy sufficiently often to the keyframe goals required for task execution. We consequently augment random keyframe sampling with specialised saliency-based keyframe selection.

We extract salient keyframes based on the acceleration profiles of selected robot bodies and the motion of the object.
The initial keyframe candidates are computed as
\begin{equation}
\mathcal{P}_{\Gamma}
= \arg\max_{\tau \in \mathcal{N}}
  \operatorname{LPF}[\Gamma](\tau).
\end{equation}
Here, $\Gamma(t)=\sum_j\|\ddot{\mathbf{p}}_j(t)\|_2$
is the aggregate acceleration signal, $\mathbf{p}_j(t)$
is body $j$'s position at frame $t$, and $\mathcal{N}$
is a local temporal neighbourhood. The low-pass filter reduces sensitivity to noise. The selected local maxima indicate rapid changes in robot motion and often correspond with transitions between task stages, such as picking or placing the object.
We further filter $\mathcal{P}_{\Gamma}$ to suppress keyframes caused by periodic locomotion.
We compute separate acceleration profiles for the left and right legs and estimate their local phase similarity and periodicity around each candidate.
Then, candidates dominated by regularly alternating leg motion are removed. This removes transient locomotion states that are unlikely to represent meaningful stages of the loco-manipulation task. We denote the resulting robot-event candidates by $\widetilde{\mathcal{P}}_{\Gamma}$.

Object motion is then used to refine the remaining candidates. We identify peaks in its vertical acceleration as object-event candidates, $\mathcal{P}_{o}$. A keyframe candidate after the object starts to move at $T_o$ is retained if it lies within a temporal tolerance $\delta$ of an object-event candidate. This gives
\begin{equation}
\mathcal{P}_{r}
=
\left\{
t \in \widetilde{\mathcal{P}}_{\Gamma}
\;\middle|\;
\exists s \in \mathcal{P}_{o},\ |t-s| \leq \delta \quad\text{or}\quad t \leq T_o
\right\}.
\end{equation}
In addition, object-event candidates that do not coincide with an existing keyframe candidate are added to ensure that important object transitions are not missed.
The salient keyframe set is therefore
\begin{equation}
\mathcal{P}_{\mathrm{salient}}
=
\{1,T\}
\cup
\mathcal{P}_{r}
\cup
\mathcal{P}_{o}^{\mathrm{new}},
\label{eq:manipulation_keyframes}
\end{equation}
where \begin{equation}
\mathcal{P}_{o}^{\mathrm{new}}
=
\left\{
s \in \mathcal{P}_{o}
\;\middle|\;
\forall t \in \mathcal{P}_{r},\ |s-t| > \delta
\right\}.
\end{equation}

\begin{figure}[t]
    \centering
    \includegraphics[width=\linewidth]{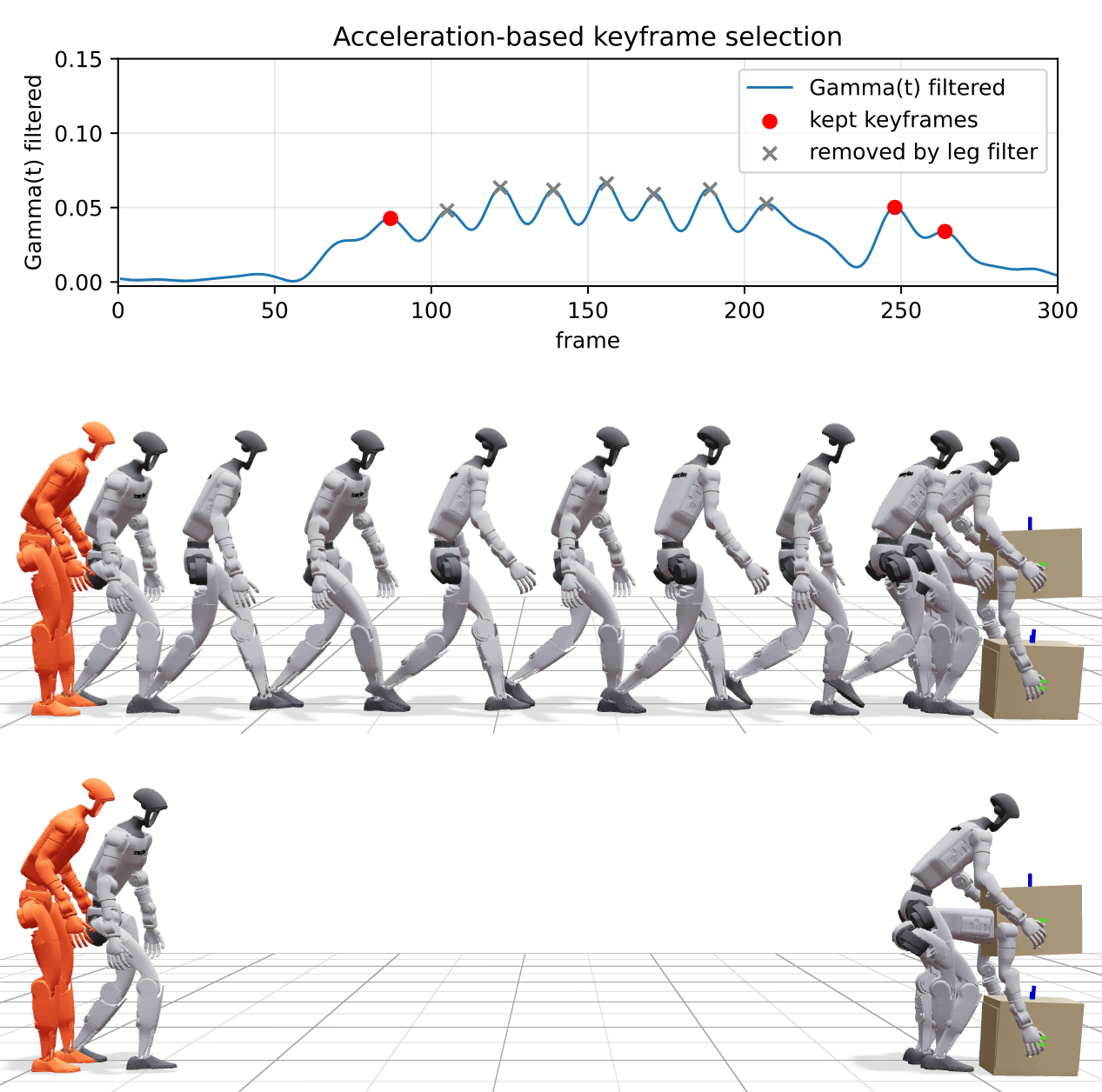}

    \caption{
    Example of the keyframe selection process. \textbf{Top:} the raw and low-pass filtered $\Gamma(t)$, where red markers indicate selected keyframes. \textbf{Middle:} selected keyframes based on acceleration ($\mathcal{P}_{\Gamma}$).
    \textbf{Bottom:} selected keyframes after leg-motion filtering ($\Tilde{\mathcal{P}}_{\Gamma}$). Most of the intermediate walking keyframes are removed while keeping salient keyframes. The orange pose is the start frame of the motion.
    }
    \label{fig:keyframe_selection_example}
    \vspace{-0.3cm}
\end{figure}

As can be seen in Fig.~\ref{fig:keyframe_selection_example}, after filtering periodic leg keyframes, the remaining set of keyframes corresponds more closely to semantically distinct stages.
This provides a better sampling strategy for policy training.

\subsubsection{Keyframe Library}
\label{sec:keyframe library}
The keyframe library includes a set of representative poses from the motion dataset. Each keyframe corresponds to a semantic motion stage in loco-manipulation, such as approaching an object, lifting, carrying, etc.
Each keyframe $\mathcal{K}_i$ contains the full pose of the robot and, if applicable, the object, specifying the nominal spatial relationship between the robot and the manipulated object.
Each keyframe is also paired with a natural language description.
During online execution, the VLM planner selects a keyframe from the library based on the text labels.
The robot and object states from the selected keyframe are then retargeted to the current scene before passing to the policy.

\subsection{Scene Alignment}

\subsubsection{Keyframe Retargeting}
\label{sec:keyframe_retargeting}

The keyframe library defines the generic robot-object configuration based on the dataset, whereas execution may involve objects with different poses and dimensions. Before a selected manipulation keyframe is passed to the low-level policy, it needs to be adapted to the target scene configuration.
Given a keyframe $\mathcal{K}$ from the library, the retargeter outputs $\mathcal{K}^{*}$ where the robot body positions are adapted to the target object using inverse kinematics.

\begin{figure}[bt]
    \centering
    \includegraphics[width=0.95\linewidth]{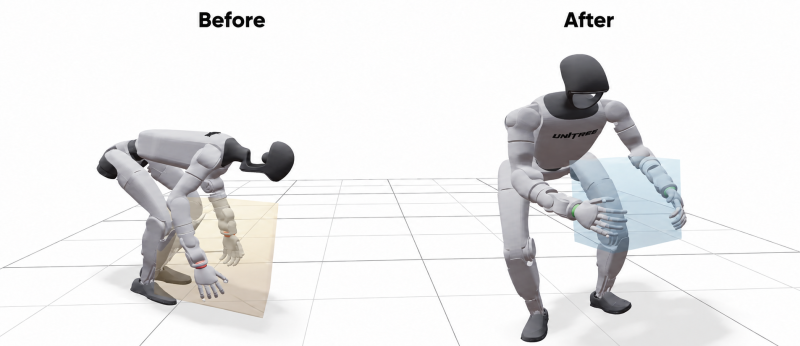}
    \caption{An example of keyframe retargeting using a general picking pose (\textbf{left}) to a different box pose and size (\textbf{right}).}
    \label{fig:new_box_retarget}
\end{figure}

First, we rotate the source object frame to align it with the observed orientation of the target object. Let the source and target object frames be
$\mathcal{B}_{i}
=
(\boldsymbol{c}_{i},\boldsymbol{R}_{i},\boldsymbol{h}_{i})$
and
$\mathcal{B}^{*}
=
(\boldsymbol{c}^{*},\boldsymbol{R}^{*},\boldsymbol{h}^{*})$,
respectively, where $\boldsymbol{c}\in\mathbb{R}^{3}$ denotes the object
centre, $\boldsymbol{R}\in SO(3)$ its orientation, and
$\boldsymbol{h}\in\mathbb{R}_{>0}^{3}$ the half-extents of its bounding box.
We then transform the end-effectors and robot root positions using
\begin{equation}
\mathcal{T}_{i\rightarrow *}(\boldsymbol{p})
=
\boldsymbol{c}^{*}
+
\boldsymbol{R}^{*}
\Big[
\big(
\boldsymbol{R}_{i}^{\top}
(\boldsymbol{p}-\boldsymbol{c}_{i})
\big)
\oslash
\boldsymbol{h}_{i}
\odot
\boldsymbol{h}^{*}
\Big],
\label{eq:box_retargeting}
\end{equation}
where $\oslash$ and $\odot$ denote element-wise division and multiplication, respectively. The transformation preserves the normalised position of each point relative to the source object while adapting it to the target object pose and dimensions. For the robot root, only the horizontal coordinates are updated.
After generating the target end-effector positions, we recover the
remaining robot configuration using iterative damped least-squares
inverse kinematics. The objective minimises the end-effector
position errors and weighted vertical foot-position residuals to keep the feet at ground height.
We use a damped least-squares solver with MuJoCo’s generalised-position integration, and the resulting coordinates are written back to the
keyframe.
The resulting
keyframe preserves the original representation while adapting the
end-effector targets and robot placement to the target object pose and
dimensions (Fig. \ref{fig:new_box_retarget}).

\subsubsection{Path Planning}
When deploying the policy, we construct a sequence of intermediate keyframes between the initial and target locomotion keyframes. This helps the robot to traverse to locations beyond the training data. The trajectory is represented by a tangent-aligned circular arc or, when necessary, the shortest valid Dubins path. Its curvature is limited to \(3\,\mathrm{m}^{-1}\), corresponding to a minimum turning radius of approximately \(0.33\,\mathrm{m}\). This constraint limits sharp turning motions not represented in the training trajectories. The path is densely sampled, after which intermediate goals are selected greedily. From each selected goal, we choose the furthest subsequent sample for which the accumulated heading change does not exceed \(45^\circ\) and the Euclidean displacement from the previous goal does not exceed 2 m.

\subsection{VLM-Based Planner}
\label{sec:vlm_planner}

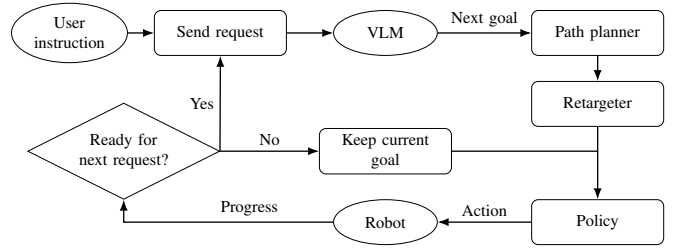
\begin{figure}[t]
    \centering
    \resizebox{\linewidth}{!}{%
    \begin{tikzpicture}[
        process/.style={
            rectangle,
            draw,
            rounded corners,
            minimum width=2.8cm,
            minimum height=0.95cm,
            align=center
        },
        decision/.style={
            diamond,
            draw,
            aspect=2,
            minimum width=3.5cm,
            minimum height=1.4cm,
            align=center
        },
        module/.style={
            ellipse,
            draw,
            minimum width=2.2cm,
            minimum height=0.95cm,
            align=center
        },
        input/.style={
            rectangle,
            draw,
            rounded corners,
            minimum width=3.0cm,
            minimum height=0.95cm,
            align=center
        },
        arrow/.style={-Latex, thick}
    ]

    \node[decision] (ready) at (-2.05, -2.5) {Ready for\\next request?};

    \node[module] (language) at (-3.2, 0) {User \\instruction};

    \node[process] (send) at (0, 0) {Send request};

    \node[module] (vlm) at (3.5, 0) {VLM};

    \node[process] (keep) at (3.5,-2.5) {Keep current\\goal};

    \node[process] (plan) at (8.0, 0) {Path planner};

    \node[process] (retarget) at (8.0,-1.5) {Retargeter};

    \node[process] (policy) at (8.0,-4) {Policy};

    \node[module] (robot) at (3.5,-4) {Robot};

    \coordinate (merge) at (8.0,-2.5);

    \draw[arrow] (ready.east) -- node[left] {Yes} (send.south);
    \draw[arrow] (language.east) -- (send.west);
    \draw[arrow] (send.east) -- (vlm.west);
    \draw[arrow] (vlm.east) -- node[above] {Next goal} (plan.west);
    \draw[arrow] (plan.south) --  (retarget.north);
    \draw[arrow] (retarget.south) -- (policy.north);

    \draw[arrow] (policy.west) -- node[above] {Action} (robot.east);

    \draw[arrow] (ready.east) -- node[above] {No} (keep.west);

    \draw[thick] (keep.east) |- (merge);

    \draw[arrow] (robot.west) ++(0,0.0) -| node[pos=0.2, above] {Progress} (ready.south);

    \end{tikzpicture}%
    }
    \caption{Overview of the VLM-based planner. The user provides a language input to the VLM. The VLM generates the next goal, which is retargeted and executed by the policy.
    When the robot reaches a stationary state, a new request is sent to the VLM. Otherwise, the system keeps the current goal and continues execution.}
    \label{fig:vlm_planner}
    \vspace{-0.4cm}
\end{figure}

The VLM-based planner provides the high-level decision-making module of the proposed hierarchy. Its role is to interpret the user instruction and the current scene, select the next semantic keyframe to be sent to the low-level policy, and decide whether the task has been completed.

The operation loop of the VLM planner is shown in Fig. \ref{fig:vlm_planner}.
At the start, the planner takes a language instruction from the user and starts the task planning loop. At each planning step, the planner constructs a request consisting of a natural-language task instruction, a visual observation of the scene, and a structured planner context. The task instruction is the long-horizon objective from user input, such as picking up a box and placing it at a target location. The visual observation is a third-person-view image that provides semantic information about the scene, while the planner context contains state and execution feedback from the robot system. The system prompt for the VLM agent describes the VLM's duties and specifies the meanings and formats for inputs and outputs.

\textbf{Planner context.}
The context is updated by the planner each time before a keyframe request is sent to the VLM server.
The context includes the previous keyframe, whether the previous goal has finished and succeeded, the current robot root pose, the current object pose, the target object pose, and controller diagnostics such as goal errors.
If the previous keyframe is successfully reached, the planner can advance to the next semantic stage; otherwise, it can retry or select a recovery keyframe.
This allows the VLM to autonomously reason about task progress and recovery.

\textbf{Constrained keyframe output.}
To make the VLM output directly compatible with the retargeter and policy, the planner is constrained to return a structured JSON response such as:
\begin{verbatim}
{ "keyframe": "crouch_to_pick_box",
  "object_to_manipulate": true,
  "task_complete": false }
\end{verbatim}

The object-manipulation flag determines whether the retargeter should preserve object-relative hand configurations and whether the object pose is treated as part of the active goal.
If the flag is false, the object-related input to the low-level policy is set to zero, and the retargeter only considers robot root position. The task-completion flag terminates the planning loop when the command has been fulfilled.

\textbf{Closed-loop execution.}
The planner is executed in an event-driven loop. A new VLM query is sent only after the robot and object have reached a stationary state and the previous goal has been given enough time to finish. This avoids querying the VLM while the low-level policy is still executing a keyframe. Once a valid VLM response is received, the selected keyframe is passed to the retargeting module which produces the goal input to the low-level policy.

\section{Results}
\label{chap:results}

To evaluate our pipeline, we focus on two types of pick-transport-place tasks: two-handed box and one-handed bucket manipulation.
The reference motion data includes 52 locomotion clips from AMASS \cite{mahmood2019amass}, 105 two-handed box manipulation clips from OmniRetarget\cite{yang2025omniretarget}, and 120 in-house recorded clips of one-handed bucket manipulation.
To improve robustness to variations in the scene, we augmented the manipulation motions by changing the initial position and orientation of the object and adjusting the robot and object trajectories accordingly~\cite{yang2025omniretarget}.

The VLM used is Qwen3.6:27B\cite{qwen36_27b} running on a single 4090 RTX GPU and interacting with the retargeter and controller through a ROS service interface.
The average time to get a keyframe output from the VLM is below 0.2s, and the mean retargeting time is around 5 ms, which results in efficient and fluent task execution.

The proposed framework is evaluated both in simulation and on hardware to assess the effectiveness of the pipeline.

\subsection{Sim-to-Sim}
\label{sec:deployment}
\begin{figure}[bt]
    \centering
    \includegraphics[width=\linewidth]{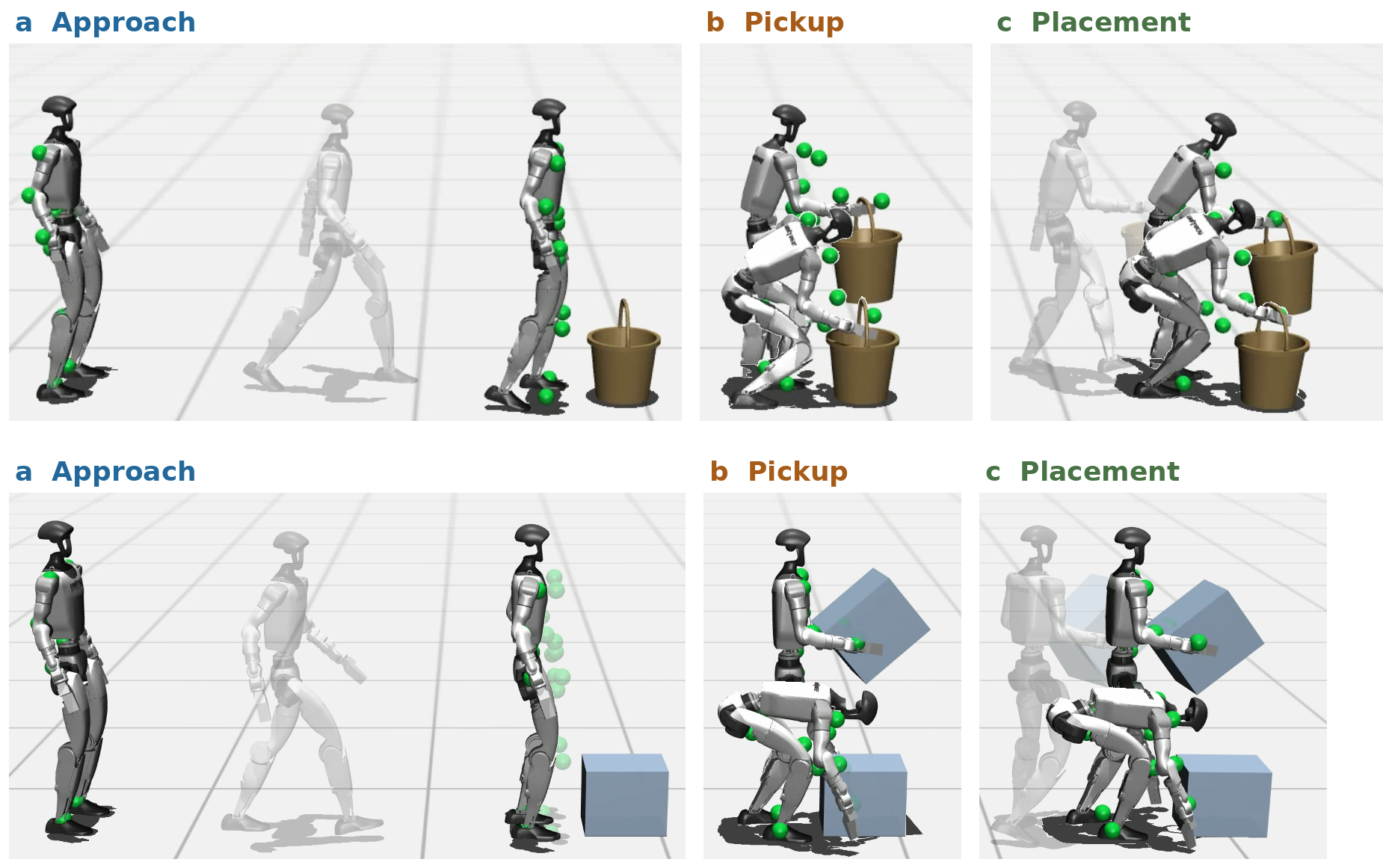}
    \caption{Qualitative results of a pick-and-place task with VLM planning for bucket (\textbf{top}) and box (\textbf{bottom}). The keyframe-conditioned policy executes a sequence of loco-manipulation behaviours. Green markers visualise the target keyframes.
    }
    \label{fig:policy_qualitative_sim}
    \vspace{-0.2cm}
\end{figure}

Sim-to-sim experiment allows us to verify the closed-loop interaction between the planner, retargeter, and policy before transferring the system to the physical robot.
As shown in Fig.~\ref{fig:policy_qualitative_sim}, our framework is able to select the next keyframe based on the task description and current robot context, and further execute the motion to reach those keyframes
The VLM is able to choose one-handed or two-handed keyframes from the library based on the asked object in the command.
\begin{figure}
    \centering
    \includegraphics[width=0.9\linewidth]{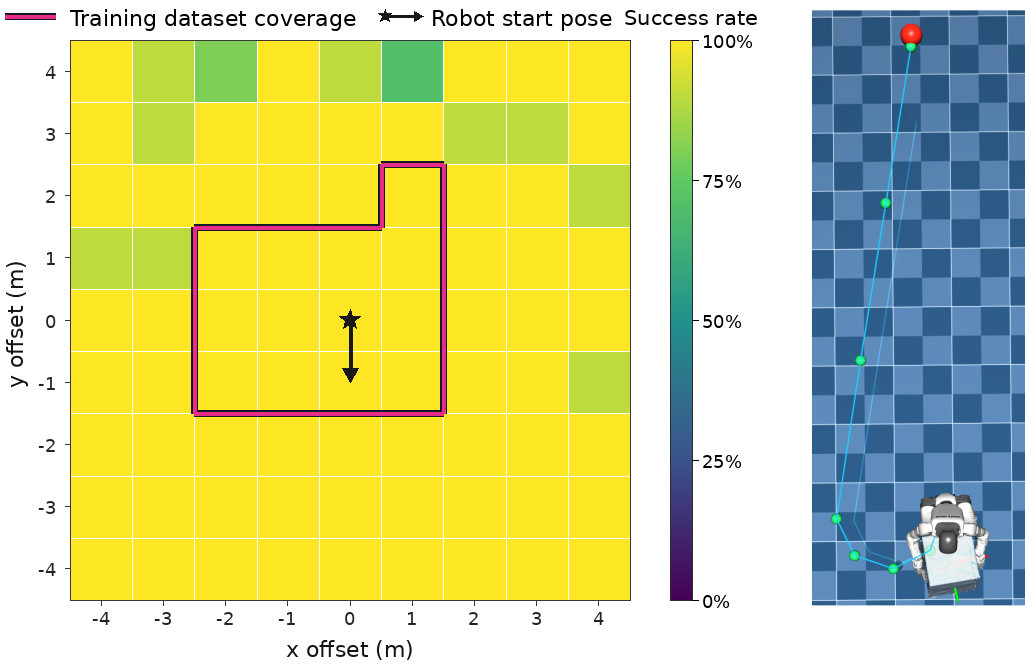}
    \caption{\textbf{Left}: Success rate for the box task with different placement positions. The red contour shows the coverage of place positions in the training data. The task is considered successful when the object is placed within 0.3 m of the target position. \textbf{Right}: Path planning for locomotion keyframes.}
    \label{fig:success rate plots}
\end{figure}

Moreover, we evaluate whether keyframe goals enable the policy to generalise beyond the loco-manipulation trajectories represented in the training dataset.
Fig.~\ref{fig:success rate plots} shows the task success rate for box pick and place.
The robot starts at the origin facing the (-y)-direction and must place the box at a specified target position.
We vary the target position over a grid and measure the policy’s success rate at each location across repeated trials in MuJoCo. A trial is considered successful if the final object position is within $0.3\mathrm{m}$ of the target. The resulting success-rate map is compared with the spatial coverage of the training data, shown by the red contour in Fig.~\ref{fig:success rate plots}. Successful placements outside this contour indicate that the policy can generalise well beyond trajectories in the reference dataset.
The figure on the right shows an example of path planning for intermediate locomotion keyframes.

\subsection{Real robot deployment}

\begin{figure*}[t]
    \centering
    \includegraphics[width=\textwidth]{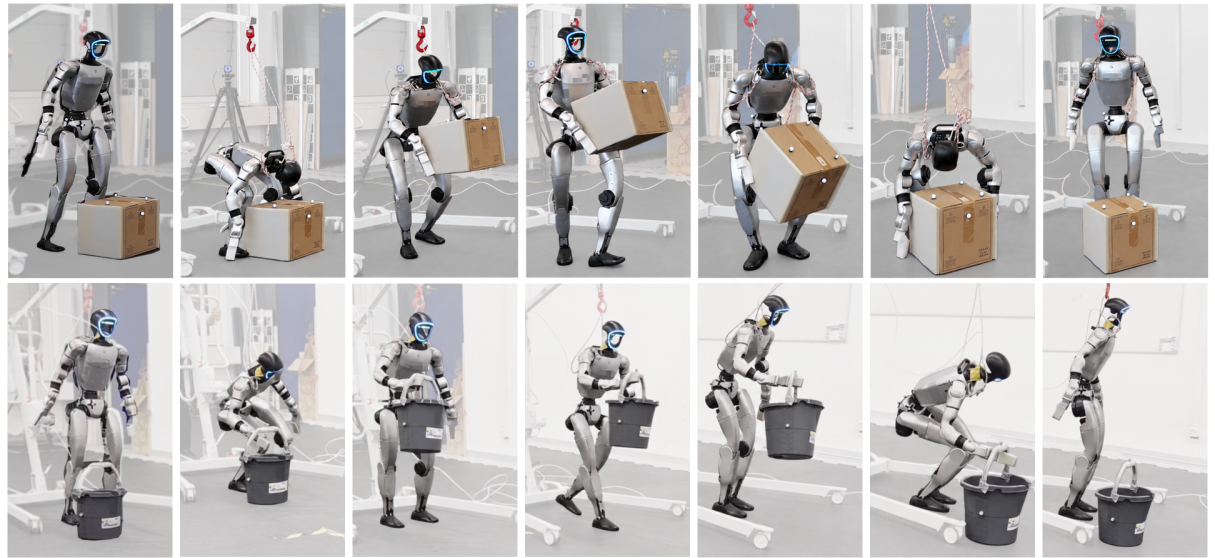}
    \caption{Hardware experiments of the full pipeline for pick-and-transport tasks with a box and a bucket. The VLM is prompted to pick and transport the object to the target, and it guides the policy to successfully complete the task.}
    \label{fig:hw}
\end{figure*}

The policy is executed on the Unitree G1 robot. The robot state is obtained from the onboard sensing, while the object relative pose is estimated externally using the motion-capture system.
Fig.
\ref{fig:hw} shows successful deployment of the policy on the robot for pick-up and placement of a box (two-handed) and a bucket (one-handed) following the keyframes chosen by a VLM.
The deployment results indicate that the proposed hierarchical system can be integrated into a real-time control pipeline, where the VLM planner, retargeter, and learned policy work together to execute long-horizon loco-manipulation behaviours with minimal planning time.

\begin{table}[t]
\centering
\caption{Comparison of success rates with 100 trials for different
task stages between policies trained with different keyframe
sampling strategies.}
\label{tab:success_rates}
\setlength{\tabcolsep}{3pt}
\resizebox{\columnwidth}{!}{%
\begin{tabular}{lcccc}
\toprule
Method & Approach & Pick & Place & End-to-end \\
\midrule
Uniform
& $97\%$ & $63\%$ & $63\%$ & $44\%$ \\
Saliency-based (ours)
& $\mathbf{99\%}$ & $\mathbf{95\%}$
& $\mathbf{93\%}$ & $\mathbf{92\%}$ \\
\bottomrule
\end{tabular}%
}
\vspace{-3pt}
\end{table}

\begin{table}[t]
\centering
\caption{Comparison of goal-reaching errors for policies
trained with uniform and saliency-based keyframe sampling.}
\label{tab:tracking_errors}
\setlength{\tabcolsep}{3pt}
\resizebox{\columnwidth}{!}{%
\begin{tabular}{lccc}
\toprule
Method & Root (m) $\downarrow$ & Joint RMSE (rad) $\downarrow$
       & Object (m) $\downarrow$ \\
\midrule
\multicolumn{4}{l}{\textbf{Approach}} \\
Uniform
& $0.125 \pm 0.050$ & $0.099 \pm 0.007$ & -- \\
Saliency-based (ours)
& $\mathbf{0.113 \pm 0.047}$ & $\mathbf{0.075 \pm 0.004}$ & -- \\
\midrule
\multicolumn{4}{l}{\textbf{Pick}} \\
Uniform
& $0.149 \pm 0.043$ & $0.150 \pm 0.020$ & $0.170 \pm 0.094$ \\
Saliency-based (ours)
& $\mathbf{0.119 \pm 0.034}$ & $\mathbf{0.116 \pm 0.007}$
& $\mathbf{0.043 \pm 0.035}$ \\
\midrule
\multicolumn{4}{l}{\textbf{Place}} \\
Uniform
& $0.162 \pm 0.048$ & $0.139 \pm 0.009$
& $\mathbf{0.082 \pm 0.026}$ \\
Saliency-based (ours)
& $\mathbf{0.067 \pm 0.018}$ & $\mathbf{0.121 \pm 0.004}$
& $0.227 \pm 0.043$ \\
\bottomrule
\end{tabular}%
}
\vspace{-3pt}
\end{table}

\subsection{Ablation on keyframe sampling}

In this section, we compare our saliency-based keyframe sampling with a uniform keyframe sampling strategy where a fixed time interval of 1.5 seconds is chosen between keyframes.
Table \ref{tab:success_rates} shows the task success rate over 100 experiments.
The starting distance to the box is varied between 0.5-2.3m and the target distance is between 0.2-2.0m of the box initial pose. The VLM is used to query the keyframes.
The task is considered successful if the final object and robot root positions error is less than 0.3m.

Table \ref{tab:tracking_errors} shows goal-reaching errors for different stages of the task.
Following an approach failure, the robot is reset near the object. Following a manipulation-keyframe failure, subsequent stages are omitted. This yields Approach/Pick/Place sample counts of 97/88/63 for uniform sampling and 99/100/93 for our method.
Our saliency-based approach achieves better reaching accuracy in most scenarios, with the sole exception of object position error during the placement phase. This is because the keyframe target for placing, taken from the reference motion, has the box too close to the robot, which makes the placing motion unstable. The RL policy corrects for this by matching the root position while placing the object further.

\subsection{Failure recovery}
\begin{figure}[bt]
    \centering
    \includegraphics[width=0.95\linewidth]{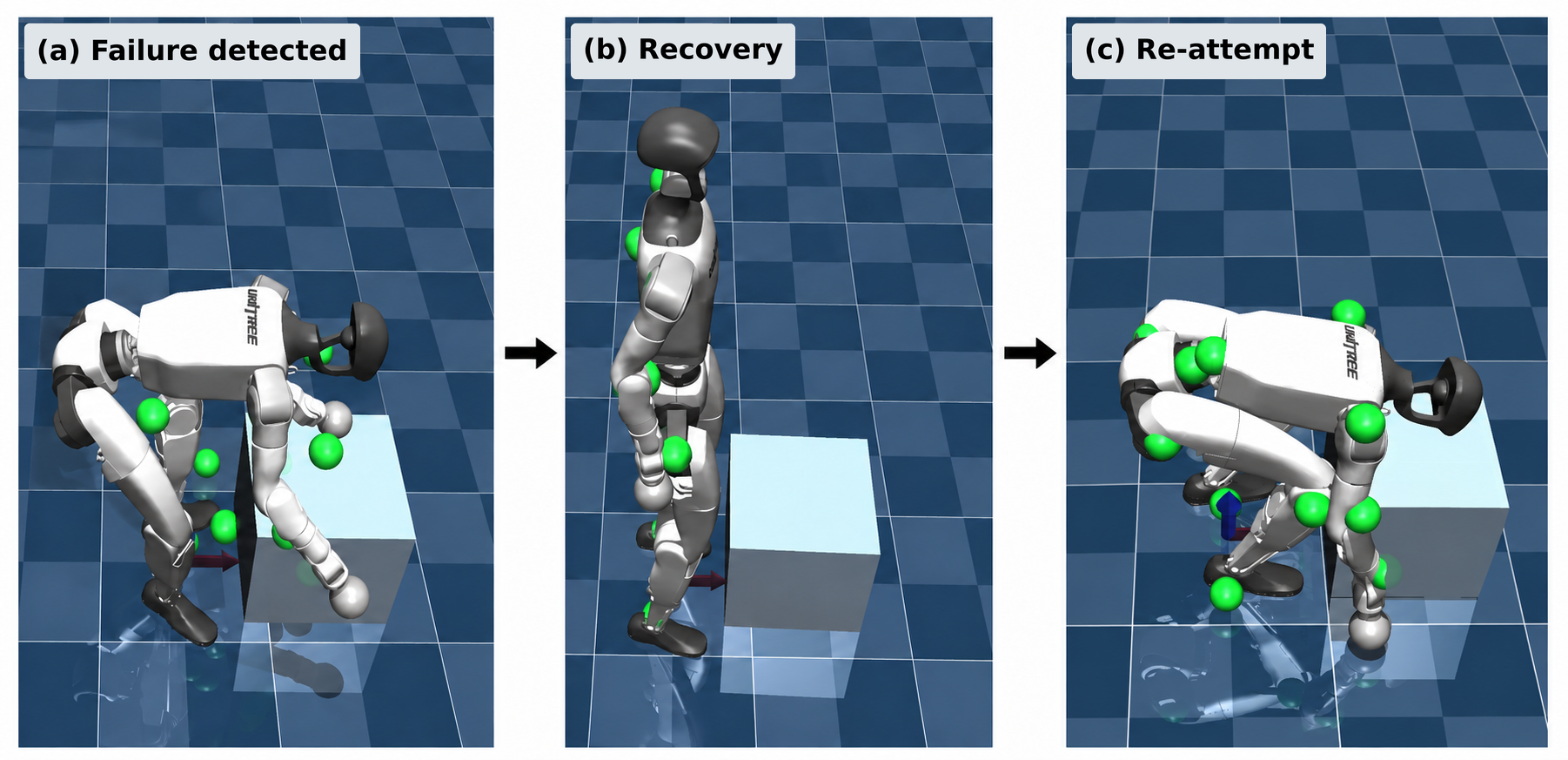}
    \caption{An example of a failure in reaching the target keyframe, and the VLM makes a recovery.}
    \label{fig:recovery_example}
    \vspace{-3pt}
\end{figure}

The VLM can also detect whether a keyframe goal is reached successfully based on the vision and errors provided in the planner context, which enables it to recover from failed attempts by resending the previous keyframe.
An example of recovery is shown in Fig. \ref{fig:recovery_example}. This demonstrates that when the policy fails to accomplish the subtask, the VLM may still guide the policy towards retrial and successful execution, emphasising the advantage of the hierarchical system.

\section{Conclusion}
This work presents KINO, a hierarchical keyframe-based framework for long-horizon humanoid loco-manipulation. By integrating the VLM planner and RL controller through motion keyframes, our system is able to execute tasks autonomously and generalise to scenarios much beyond the training data. With the saliency-based keyframe selection, the task success rate increased from 44\% to 92\% compared to the uniform baseline.  The simulation and hardware experiments demonstrated that the VLM is able to plan for different objects and choose the appropriate one- or two-handed manipulation keyframes. It can also detect failures and re-attempt certain motions autonomously, making the system highly generalisable and robust.

Despite these advantages, some limitations remain. The current keyframe library is manually constructed and labelled, which limits scalability as the number of tasks
increases. A promising direction is to automatically extract salient keyframes and use VLMs to assign semantic descriptions to construct a reusable library with less manual
effort. Furthermore, the current system relies on third-view visual observations for the planner and external mocap measurements for relative object pose. In-the-wild deployment
will require replacing these with onboard perception, such as egocentric RGB-D sensing.

Overall, the results suggest that motion keyframes provide a practical and interpretable interface between semantic planning and whole-body control.
Scaling the framework to more diverse tasks and operating using only onboard sensing would be an important step towards general-purpose humanoid manipulation in real-world environments.

\bibliographystyle{ieeetr}
\bibliography{references}
\end{document}